\documentclass{article}
\usepackage{spconf,amsmath,graphicx}

\usepackage{enumitem}
\setlist{nosep, leftmargin=14pt}

\usepackage{mwe} 
\usepackage{amssymb}
\usepackage{stmaryrd}

\usepackage{ascmac}
\usepackage{booktabs}
\usepackage{multirow}
\usepackage{multicol}

\usepackage{pifont}
\usepackage[table]{xcolor}
\usepackage{siunitx}
\usepackage{caption} 
\usepackage{cuted} 
\usepackage{cite}

\title{VesselFusion: Diffusion Models for Vessel centerline extraction from 3D CT Images}
\name{Soichi Mita$^{1}$ \qquad Shumpei Takezaki$^{1,2}$ \qquad Ryoma Bise$^{1}$}
\address{$^1$Kyushu University, Fukuoka, Japan \\ $^2$German Research Center for Artificial Intelligence (DFKI), Kaiserslautern, Germany}

\begin{document}
\ninept
\maketitle

\begin{abstract}
Vessel centerline extraction from 3D CT images is an important task because it reduces annotation effort to build a model that estimates a vessel structure. It is challenging to estimate  natural vessel structures since conventional approaches are deterministic models, which cannot capture a complex human structure. In this study, we propose VesselFusion, which is a diffusion model to extract the vessel centerline from 3D CT image. The proposed method uses a coarse-to-fine representation of the centerline and a voting-based aggregation for a natural and stable extraction. VesselFusion was evaluated on a publicly available CT image dataset and achieved higher extraction accuracy and a more natural result than conventional approaches. 
\end{abstract}


\section{Introduction}
\label{sec:intro}


Vessel extraction has played a crucial role in providing detailed structural information from 3D CT images to support surgical planning, diagnosis, and computational hemodynamics~\cite{Lesage2009-intro1, Wolterink2018centerline_tracing,He2020-intro1,Salahuddin2020-intro1,Chen2024-intro1}. Especially, deep learning-based models have achieved high-quality vessel segmentation~\cite{Tetteh2020deepvesselnet,Yagis2024vascularsegmentation}; however, these methods need accurate annotation of vessel segmentations. It is challenging to create dense 3D annotations because it is labor-intensive and ambiguous, as vessel boundaries in volumetric data are often unclear, introducing annotation noise and inconsistency. 

To remedy this challenge, vessel centerline extraction has attracted attention. 
A vessel centerline is a set of coordinates describing the vessel tree; this compact representation reduces annotation effort and is less affected by boundary ambiguity while still capturing vessel structures~\cite{Zhao2024,prabhakar2023vesselformer}.
However, existing centerline-based methods typically rely on deterministic regression or heuristic tracing, making them sensitive to noise and prone to broken or spurious vessel segments~\cite{Wolterink2018centerline_tracing,He2020-intro1, Naeem2024trexplorer, Naeem2025trexplorersuper}.

\begin{figure}[t]
    \centering
    \includegraphics[width=0.95\linewidth]{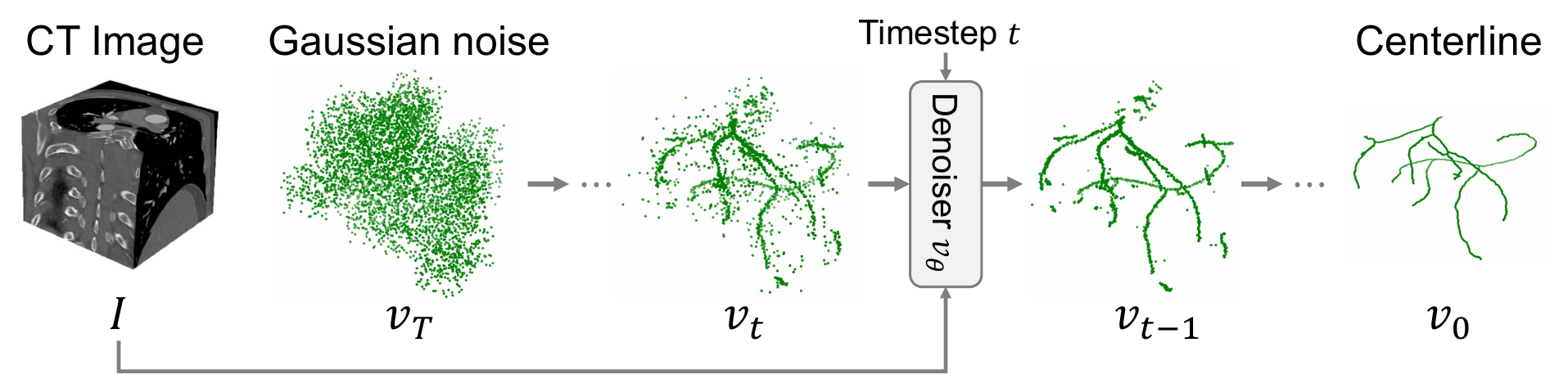}
    \caption{Overview of \textbf{VesselFusion}: A conditional diffusion model extracts a vessel centerline from a 3D CT image.}
    \label{fig:VesselDiff}
\end{figure}

\begin{figure}[t]
    \centering
    \includegraphics[width=0.95\linewidth]{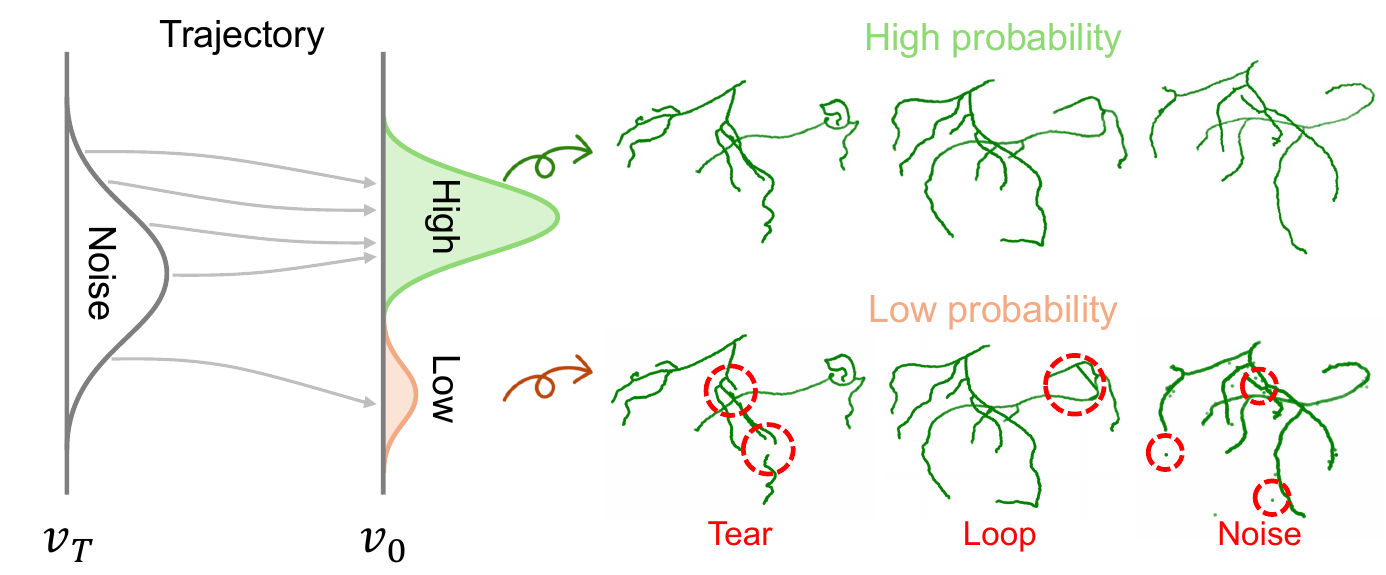}\\[-2mm]
    \caption{Stochastic generation from Gaussian distribution to a vessel distribution. VesselFusion infrequently generates unnatural structure~(e.g., tears, loops, noise).}
    \label{fig:GenerativeModel}
    \vspace{-10pt}
\end{figure}

In this study, we propose \textbf{VesselFusion} as shown in Fig.~\ref{fig:VesselDiff}, a conditional diffusion model that generates vessel centerlines conditioned on CT images. As shown in Fig.~\ref{fig:GenerativeModel}, VesselFusion can generate more natural vessel structures than deterministic methods because its learned distributions capture the variability of natural vessel shapes while suppressing unlikely ones.
This method also uses a coarse-to-fine (C2F) representation, which encodes coordinates in a structured form that combines discrete and continuous spatial components, instead of the original vessel-centerline coordinates, because this C2F representation is more tractable within a diffusion-based framework. Moreover, since VesselFusion may sometimes produce unnatural structures from a single stochastic generation, as shown in Fig.~\ref{fig:GenerativeModel}, we apply voting-based aggregation that averages multiple predictions generated from different initial noises to achieve more stable results.

We evaluated VesselFusion on a clinical CT dataset~\cite{Zeng2023imagecas} and compared it with existing centerline extraction methods. The results showed that VesselFusion achieves higher extraction accuracy and produces more natural vessel structures.
Our main contributions are summarized as follows:
\begin{itemize}
    \item We propose VesselFusion, the first method using a generative model for the vessel centerline extraction from CT images.
    \item We apply coarse-to-fine representation and voting-based estimation to estimate a more natural structure.
    \item Experimental evaluations with a public CT image dataset confirm our proposed method outperforms conventional methods in estimation accuracy and naturalness of vessel structure.
\end{itemize}

\section{Related Work}
\label{sec:related}

\noindent
{\bf Vessel centerline extraction:}
Conventional methods of vessel centerline extraction have used two approaches. One type of conventional approach has applied iterative tracking structures from initial coordinates~\cite{Wolterink2018centerline_tracing,Salahuddin2020-intro1,Naeem2024trexplorer,Naeem2025trexplorersuper,Bise2016trace}. While this approach estimates smooth and continuous structures, it has limited applicability in clinical environments because it requires initialization of coordinates~\cite{Naeem2025trexplorersuper}. Another approach has applied graph representations of vessel structures. This approach predicts topology directly from CT images~\cite{prabhakar2023vesselformer,Shit2022relationformer}. For example, VesselFormer~\cite{prabhakar2023vesselformer} formulates estimation of the centerline as an end-to-end graph generation problem, eliminating initialization of coordinates, and can explicitly model bifurcations. However, because this approach uses deterministic estimation, it often causes structural inconsistencies such as missing or spurious connections. Compared with these two approaches, VesselFusion can estimate natural structures without initialization because our method leverages a generative model for the centerline extraction.

\noindent
{\bf Diffusion models in medical domain:}
In the medical domain, diffusion models have achieved strong performance in many tasks, such as segmentation, detection, and generation~\cite{Kazerouni2023diffsurvey,Shi2025diffsurvey,Luo2025diffsurvey}. Especially, in the segmentation task, diffusion models have been a critical approach~\cite{Wu2022MedSegDiff,Wu2024MedSegDiffV2}. Deterministic approach often may estimate unnatural structure because human anatomy is complex. In contrast, the diffusion models can estimate natural ones because the generative models learn the distribution of human anatomy. To the best of our knowledge, VesselFusion is the first method to perform vessel centerline extraction from 3D CT images.

\section{VesselFusion}
\label{sec:methods}

\subsection{Overview}
We propose VesselFusion, a conditional diffusion model for vessel centerline extraction from CT images, as shown in Fig.~\ref{fig:VesselDiff}.
Unlike voxel-wise segmentation methods, VesselFusion directly estimates vessel groups of coordinates representing centerlines conditioned on CT images. Let the training dataset consist of CT images $\mathcal{I} = \{ I_i \}_{i=1}^{N}$ and their manually traced set of vessel centerline coordinates $\mathcal{P} = \{ P_i \}_{i=1}^{N}$, where $P_i = \{(x_i^j, y_i^j, z_i^j)\}_{j=1}^{M_i}$.  
Our goal is that, given an unknown CT image \(I\), VesselFusion predicts a realistic set of vessel centerline coordinates\(\bar{P}\).

\begin{figure}
    \centering
    \includegraphics[width=0.95\linewidth]{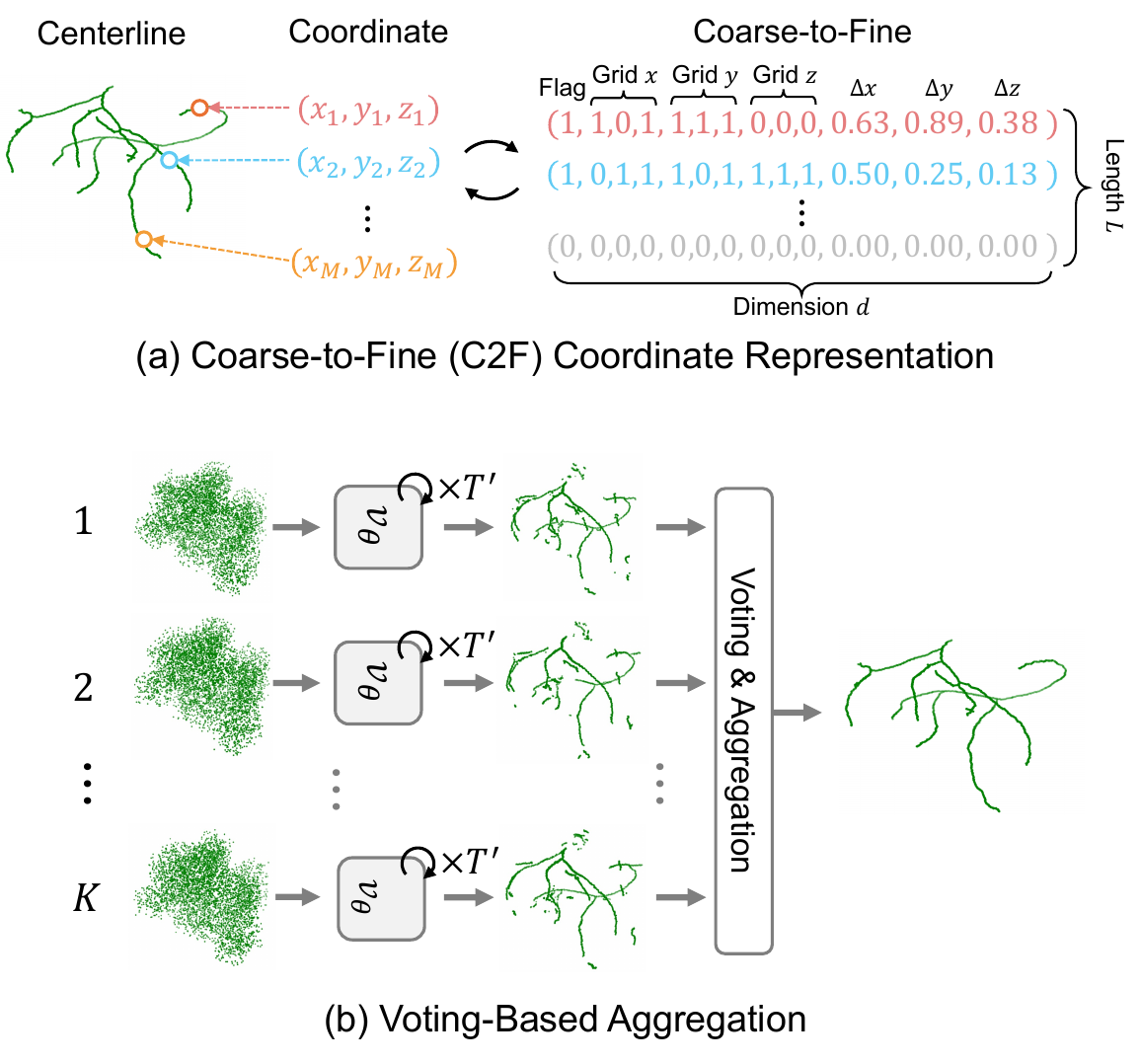}\\[-2mm]
    \caption{(a)~Representation of  centerline coordinates by a coarse-to-fine style. (b)~Voting and aggregation of some estimations from different initial noises.}
    \label{fig:C2F}
    \vspace{-10pt}
\end{figure}

We also apply two components to standard diffusion models for building VesselFusion. The first component is a coarse-to-fine representation of the vessel centerline as illustrated in Fig.~\ref{fig:C2F}~(a). Each centerline coordinate is encoded by a combination of a coarse grid index and a local offset, forming a structured coordinate representation suitable for diffusion-based generation. The second component is voting-based aggregation to estimate natural structure as shown by Fig.~\ref{fig:C2F}~(b). 
At a single estimation, VesselFusion may estimate unnatural structure. Therefore, we integrate multiple estimations from different initial noises to obtain a stable and anatomically consistent centerline.

\subsection{Coarse-to-Fine Coordinate Representation}
As shown in Fig.~\ref{fig:C2F}~(a), we apply a coarse-to-fine~(C2F) coordinate representation inspired by VecFusion~\cite{Thamizharasan2023vecfusion} because directly representing original 3D coordinates within diffusion models is unstable due to data sparsity and scale imbalance. The C2F representation encodes sets of the centerline coordinates in a structured form that combines discrete and continuous spatial components.

Specifically, given a CT image $ I\in \mathbb{R}^{V_x \times V_y \times V_z}$, the 3D coordinate is uniformly divided into $G_x \times G_y \times G_z$ grid cells along the \(x\)-, \(y\)-, and \(z\)-axes. Each element $p=(x,y,z)\in P$ at a vessel centerline is represented by:  

\noindent
\textbf{Flag} \(f \in \{0,1\}\) indicates whether the entry is valid ($1 =$ valid, $0 =$ padding), allowing variable-length sets to be represented by a fixed-size representation.  

\noindent
\textbf{Grid-cell indices} \((g_x,g_y,g_z)\) denote the indices of the grid cell containing \(p\), computed as  
\[
g_x = \left\lfloor \frac{x}{s_x} \right\rfloor,\quad
g_y = \left\lfloor \frac{y}{s_y} \right\rfloor,\quad
g_z = \left\lfloor \frac{z}{s_z} \right\rfloor,
\]
where \(s_x=V_x/G_x\), \(s_y=V_y/G_y\), and \(s_z=V_z/G_z\) are the grid-cell sizes along each axis. 

Moreover, we convert each grid-cell index into a binary vector using an encoding function $\phi_\mathrm{bin}: \mathbb{N}\rightarrow\{0,1\}^{B}$, where $B=\log_2 G$. Applying this function to an index $g$ yields its binary representation $b$.
\[
b = (b^1,b^2,\dots,b^B) = \phi_\mathrm{bin}(g),
\]
where each digit is converted into a binary vector corresponding to each dimension. We denote the binary vector corresponding to each axis and each element, such as $b_x = (b_{x}^1,b_{x}^2,\dots,b_{x}^B)$.

\noindent
\textbf{Local offsets} \((\Delta x,\Delta y,\Delta z)\) represent the relative position from the grid-cell center \((h_x,h_y,h_z)=(s_x(g_x + \frac{1}{2}),s_y(g_y + \frac{1}{2}),s_z(g_z + \frac{1}{2}))\):  
\[
\Delta x = x - h_x,\quad
\Delta y = y - h_y,\quad
\Delta z = z - h_z.
\]
Each offset is then normalized by the grid-cell size so that its magnitude is scaled within a consistent range.

Finally, each centerline element is expressed as
\[
c = (f, b_{x}^1,b_{x}^2,\dots,b_{z}^{B},\Delta x,\Delta y,\Delta z),
\]
and the set of all centerline elements is denoted as
\(\mathcal{C} = \{c_j\}_{j=1}^{M}\), where $c_j$ is a C2F representation of $j$-th element.  
To handle a fixed vector format for a diffusion model, \(\mathcal{C}\) is stacked into a matrix and are padded with zeros of all components to reach a maximum length. Finally, we obtain a vector of C2F representation $v_0 \in \mathbb{R}^{L \times d}$, where $L$ is the maximum length for fixed format and $d$ is the dimension of C2F representations.

\subsection{CT-Conditioned Diffusion Model}
We use the denoising diffusion probability model~\cite{Ho2020ddpm} conditioned on the CT image to estimate the vessel centerline. Given a C2F representation \(v_0\) and the corresponding CT image \(I\), the diffusion process gradually perturbs \(v_0\) by adding Gaussian noise as
\[
v_t \;=\; \sqrt{\gamma(t)}\,v_0 \;+\; \sqrt{1-\gamma(t)}\,\epsilon,
\quad \epsilon \sim \mathcal{N}(0,1),\;\; t\in[0,T],
\]
where \(\gamma(t)\in[0,1]\) is monotonically decreasing noise scheduler: $\gamma(0)=1 > \cdots >  \gamma(t) > \gamma(t+1) > \cdots > \gamma(T)=0$, and $T$ is a diffusion timestep in the training.

During training, the denoiser $v_\theta$, where $\theta$ denotes the trainable parameters, is optimized by minimizing the mean squared error between the predicted and true C2F representation:
\[
\mathcal{L}_{\mathrm{diff}} \;=\; \mathbb{E}_{t\sim\mathcal{U}(1,T)}\left[ \big\| v_\theta(v_t,I, t) - v_0 \big\|^2\right ],
\]
where $\mathcal{U}(1,T)$ is a uniform distribution from $1$ to $T$.
Through the training using the loss, $v_\theta$ learns to estimate the  C2F representation conditioned on CT images at each timestep~$t$.

At the inference, we apply the DDIM sampling~\cite{Song2020ddim}, which deterministically reverses the process from $T$ to $0$. Given random initial noise \(\bar{v}_{T} \sim \mathcal{N}(0,1)\), this sampling uses iterative denoising, defined as:
\[
\bar{v}_{t-\Delta t} \;=\; \sqrt{\gamma({t-\Delta t})}\,\hat{v}_0(t) \;+\; \sqrt{1-\gamma(t-\Delta t})\ \hat{\epsilon}(t),
\]
where $\hat{v_0}(t)$ and $\hat{\epsilon}(t)$ are predicted the $v_0$ and $\epsilon$ at timestep $t$,
\[
\hat{v_0}(t) = v_\theta(\bar{v}_t, I, t),\ \ \hat{\epsilon}(t) = \frac{\,v_t - \sqrt{\gamma(t)}\,{v}_\theta(\bar{v}_t,I,t)\,}{\sqrt{1-\gamma(t)}}.
\]
Here, $\Delta t$ denotes how many timesteps are skipped during inference.
If we use an inference timestep $T' \leq T$, we define $\Delta t=T/T'$. 

When we obtain $\bar{v}_{0}$ using the DDIM sampling, $\bar{v}_{0}$ is converted into estimated coordinates $\bar{P}$. First, we remove the elements for which $f=0$ in $\bar{v}_{0}$, denoting $\bar{v}_\mathrm{valid}$. Second, a thresholding process is applied to $\bar{v}_\mathrm{valid}$. Specifically, we apply a threshold value $\lambda$ for elements of binary vectors $b$ of each axis grid-cells in $\bar{v}_\mathrm{valid}$, setting the value of each dimension to 1 if it is greater than $\lambda$, and to 0 otherwise. Then, we can obtain \(\bar{\mathcal{C}} = \{\bar{c}_j\}_{j=1}^{\bar{M}}\), where $\bar{M} \leq L$ is a valid length in the estimated representations. Finally, we convert \(\bar{\mathcal{C}}\) into \(\bar{P}=\{(\bar{x}^j,\bar{y}^j,\bar{z}^j)\}_{j=1}^{\bar{M}}\), where each axis coordinate is calculated as:
\[
\bar{x} = \left\lfloor s_x \times (\phi_\mathrm{bin}^{-1}(\bar{b}_x) + \Delta \bar{x}) + h_x \right\rceil,
\]
where $\left\lfloor \cdot \right\rceil$ means a round operation, $\bar{b}_x$ and $\Delta \bar{x}$ are the estimated binary vector of the grid-cell and the local offset on the $x$-coordinate. The $y$- and $z$- coordinates follow the same formulation.



\subsection{Voting-Based Aggregation}
For stable estimation, we introduce voting-based aggregation as shown in Fig.~\ref{fig:C2F}~(b), which is the estimation that averages multiple inferences from different noise. The DDIM sampling is even stochastic due to the randomness of the initial noise. It causes the estimation of unnatural structures, which include loops, tears, and noise. To overcome the problem, we aggregate multiple estimated sets of coordinates through voxel-wise voting to obtain a stable and anatomically faithful estimate of the vessel structure.  

Specifically, we generate \(K\) sets of the coordinates, which are denoted as $\bar{\mathcal{P}}=\{\bar{P}_k\}_{k=1}^K$, and perform voting in a coordinate space. We define a discrete coordinate space:
\[
\Omega:=\{{0,\dots,V_x-1}\} \times \{{0,\dots,V_y-1}\} \times \{{0,\dots,V_z-1}\}
\]
For voting of coordinates at $\Omega$, the voting function 
$\psi_\mathrm{vote}: \Omega \rightarrow \mathbb{N}_{\geq 0}$ is defined as below:
\[
\psi_\mathrm{vote}(x,y,z) = \sum_{k=1}^{K} \llbracket (x,y,z) \in \bar{P}_k \rrbracket,
\]
where $\llbracket \cdot \rrbracket$ is Iverson bracket, which evaluates to 1 if the predicate is true and 0 otherwise. We aggregate this voting and then can obtain the voting estimate:
\[
\bar{P}_\mathrm{vote}^{\tau} = \{(x,y,z)\in \Omega\mid\psi_\mathrm{vote}(x,y,z)\geq \tau\},
\]
where $\tau$ is a voting threshold value. Moreover, the voting threshold value $\tau$ can automatically be calculated as
\[
\tau^{*}
= \min_{\tau\in\{1,2,\dots,K\}}\ \left\lvert  \frac{1}{K}\sum_{k=1}^{K} |\bar{P}_k| - |\bar{P}_\mathrm{vote}^{\tau} |\right\rvert.
\]
The threshold value $\tau^{*}$ brings closer the number of elements for voting estimation $\bar{P}_\mathrm{vote}^{\tau}$ to the average of each set $\bar{P}_k$.  
\begin{table*}[t]
\centering
\caption{Comparison of vessel extraction accuracy (Prec., Rec., F1 for different radii $R$) and structural consistency (Betti numbers).}
\label{tab:quant_results}
\vspace{-2mm}
\small
\setlength{\tabcolsep}{4pt} 
\begin{tabular*}{\textwidth}{@{\extracolsep{\fill}} l ccc ccc ccc cc}
\toprule
\multirow{2}{*}{Method} &
\multicolumn{3}{c}{$R=1$} &
\multicolumn{3}{c}{$R=2$} &
\multicolumn{3}{c}{$R=3$} &
\multicolumn{2}{c}{Betti} \\
\cmidrule(lr){2-4}\cmidrule(lr){5-7}\cmidrule(lr){8-10}\cmidrule(lr){11-12}
& Prec. $\uparrow$ & Rec. $\uparrow$ & F1 $\uparrow$ & Prec. $\uparrow$ & Rec. $\uparrow$ & F1 $\uparrow$ & Prec. $\uparrow$ & Rec. $\uparrow$ & F1 $\uparrow$ & Betti-0 $\downarrow$ & Betti-1 $\downarrow$ \\
\midrule
Baseline &
0.838 & 0.651 & 0.733 &
0.851 & 0.789 & 0.819 &
0.856 & 0.829 & 0.842 &
190.66 & \textbf{0.04} \\
VesselFormer &
0.651 & 0.616 & 0.633 &
0.739 & \textbf{0.806} & 0.771 &
0.776 & \textbf{0.879} & 0.824 &
68.29 & 4.51 \\
VesselFusion &
\textbf{0.855} & \textbf{0.679} & \textbf{0.757} &
\textbf{0.880} & 0.786 & \textbf{0.830} &
\textbf{0.887} & 0.830 & \textbf{0.858} &
\textbf{67.66} & 0.09 \\
\bottomrule
\end{tabular*}
\end{table*}

\begin{table*}[t]
\centering
\caption{Ablation study on the effects of coarse-to-fine representation (C2F) and voting-based aggregation. 
Higher values ($\uparrow$) indicate better accuracy, while lower values ($\downarrow$) indicate better structural consistency.}
\label{tab:ablation_results}
\vspace{-2mm}

\small
\setlength{\tabcolsep}{4pt}
\begin{tabular*}{\textwidth}{@{\extracolsep{\fill}} ll ccc ccc ccc cc}
\toprule
\multirow{2}{*}{C2F} & \multirow{2}{*}{Vote} &
\multicolumn{3}{c}{$R=1$} &
\multicolumn{3}{c}{$R=2$} &
\multicolumn{3}{c}{$R=3$} &
\multicolumn{2}{c}{Betti} \\ 
\cmidrule(lr){3-5}\cmidrule(lr){6-8}\cmidrule(lr){9-11}\cmidrule(lr){12-13}
& & Prec. $\uparrow$& Rec. $\uparrow$& F1 $\uparrow$& Prec. $\uparrow$& Rec. $\uparrow$& F1 $\uparrow$& Prec. $\uparrow$& Rec. $\uparrow$& F1 $\uparrow$& Betti-0 $\downarrow$ & Betti-1 $\downarrow$ \\
\midrule
   &            & \num{0.459406} & \num{0.299848} & \num{0.362861}
                & \num{0.624654} & \num{0.605313} & \num{0.614832}
                & \num{0.702071} & \num{0.781351} & \num{0.739592}
                  & 509.35 & 2.53 \\
\multicolumn{1}{c}{\checkmark} & 
                  & \num{0.638089} & \num{0.612302} & \num{0.624930}
                  & \num{0.693412} & \textbf{0.806} & \num{0.745560}
                  & \num{0.717571} & \textbf{0.867} & \num{0.785346}
                  & 115.06 & 3.16 \\
   & \multicolumn{1}{c}{\checkmark} &
                  \num{0.679408} & \num{0.358461} & \num{0.469310} &
                  \num{0.842009} & \num{0.582883} & \num{0.688884} &
                  \textbf{0.893} & \num{0.715583} & \num{0.794350} &
                  105.65 & \textbf{0.07} \\
\multicolumn{1}{c}{\checkmark} & \multicolumn{1}{c}{\checkmark}
                  & \textbf{0.855} & \textbf{0.679} & \textbf{0.757}
                  & \textbf{0.880} & \num{0.785814} & \textbf{0.830}
                  & \num{0.887398} & \num{0.829663} & \textbf{0.858}
                  & \textbf{67.66} & 0.09 \\
\bottomrule
\vspace{-20pt}
\end{tabular*}

\end{table*}

\section{Experimental Results}
\subsection{Experimental Setup}
\label{sec:Results}
\noindent\textbf{Dataset:}
We used the public ImageCAS dataset~\cite{Zeng2023imagecas} of 1,000 coronary CT volumes with vessel segmentation masks (512×512 in-plane, 206–275 slices). Vessel centerlines were obtained from the masks by 3D skeletonization.
The dataset was split into training/validation/test in a 7:1:2 ratio.

\noindent\textbf{Implementation details:}
The denoiser was based on a Transformer encoder~\cite{Devlin2018transformer} that predicts clean C2F representations from noisy inputs, conditioned on CT features extracted by a lightweight 3D CNN.
The maximum number of vessel elements was set to \(L=320\) and the grid size to \(G_x=G_y=G_z=16\).  
Training was conducted for 300 epochs with a batch size of 48, using AdamW~\cite{Loshchilov2017adamw} (learning rate 0.0001, weight decay 0.01).  
The model at the epoch achieving the best validation performance was used for inference. 
During inference, we used $T'=100$ DDIM sampling steps. 
Following AnalogBits~\cite{Chen2022analogbits}, we optimize the binary-encoded dimensions as continuous variables, and round these dimensions at inference 
to recover their binary values, while the offset dimensions remain continuous.
We generated $K=100$ samples from different random noise initializations for 
voting-based aggregation.

\noindent\textbf{Comparison methods:} 
To demonstrate the effectiveness of VesselFusion, we compared the accuracy of extracted vessels with the following methods: 
(1) a U-Net~\cite{Ronneberger2015unet} trained with binary masks that treat the vessel centerline as foreground,  
(2) VesselFormer~\cite{prabhakar2023vesselformer}, which predicts vessel graphs from 3D CT images.

\noindent\textbf{Evaluation metrics:} 
For quantitative evaluation, we follow the coordinate-based evaluation used in prior work~\cite{Schaap2009eval,Yamane2025centerline_extraction}. 
Precision and recall are computed using a distance threshold $R$ with different matching directions. 
For precision, a predicted point is counted as a true positive if any ground-truth point lies within $R$. 
For recall, the same rule is applied in the opposite direction (ground truth to prediction). 
Unmatched predicted points become false positives, and unmatched ground-truth points become false negatives. 
We evaluated the results at $R = 1, 2, 3$; a higher score at $R = 1$ indicates a more precise estimation of vessel coordinates.
To further assess the structural plausibility of the extracted vessels, we evaluated the number of disconnected components (Betti-0) and loops (Betti-1).
Fragmented centerlines or noise-like artifacts increase Betti-0, while spurious loop formation increases Betti-1.
As the ground-truth vessel centerlines are loop-free and highly connected, accurate vessel inference should yield low Betti-0 and Betti-1 values, indicating anatomically realistic vessel structures~\cite{Naeem2025trexplorersuper}.


\begin{figure}[t]
    \centering
    \includegraphics[width=\linewidth]{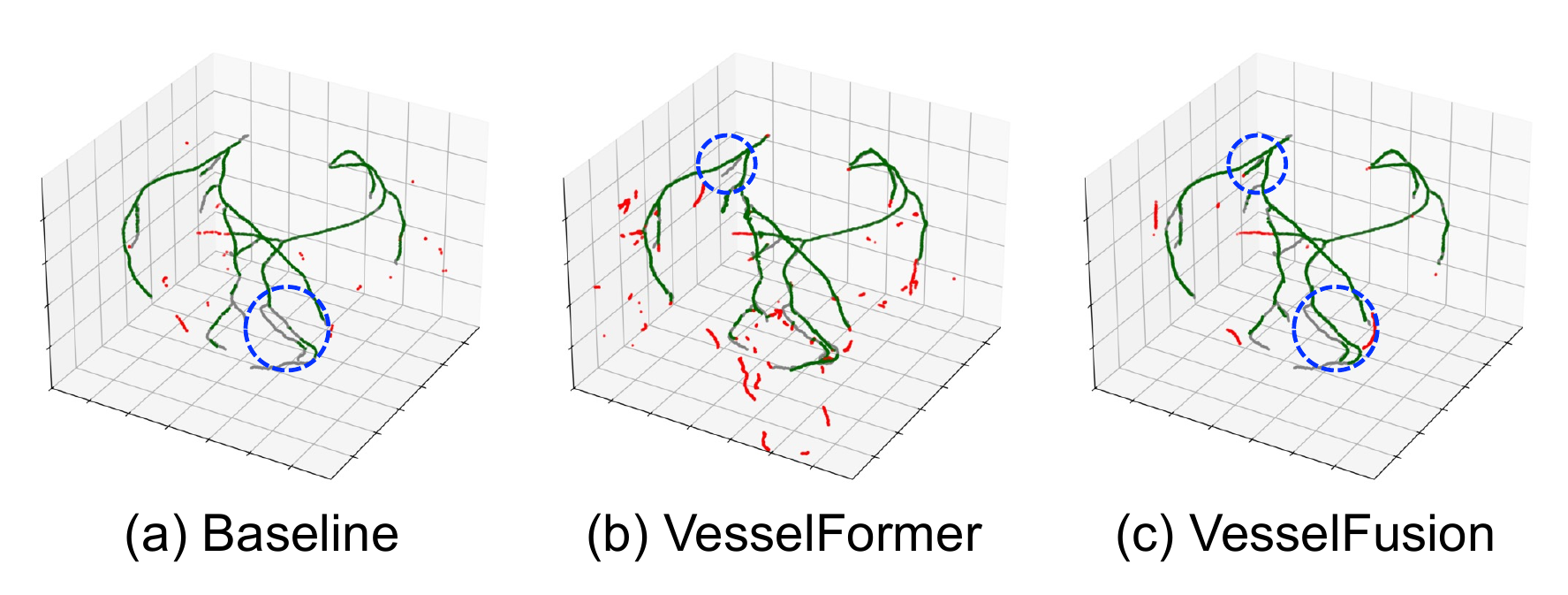}\\[-2mm]
    \caption{Qualitative comparison of vessel centerline extraction: (a)~Baseline U-Net, (b)~VesselFormer, (c)~VesselFusion. \textcolor[HTML]{3B7D23}{Green}: True positive (correctly extracted vessels). \textcolor{red}{Red}: False positive. \textcolor{gray}{Gray}: Ground truth. \textcolor{blue}{Blue}: Areas that were missed by the comparison method but were correctly detected by VesselFusion.}
    \label{fig:ExtractedVessels}
    \vspace{-10pt}
\end{figure}

\subsection{Experimental Results}
The quantitative results are summarized in Table~1.  
VesselFusion achieved the highest F1-scores for all \(R\), demonstrating superior vessel extraction accuracy.  
In particular, the improvement at \(R=1\) indicates that our model can predict vessel coordinates with higher spatial precision.Regarding structural evaluation, the baseline is very poor for Betti-0, and VesselFormer improves on Betti-0, but Betti-1 is high. VesselFusion improves both Betti-0 and Betti-1 simultaneously, demonstrating a natural estimation of the vessel structure.

As shown in Fig.~\ref{fig:ExtractedVessels}, the baseline and VesselFormer produce many noisy predictions that are anatomically implausible, whereas VesselFusion successfully suppresses such artifacts. This suggests that VesselFusion performs inference while preserving the vessel structure, which is thought to improve coordinate accuracy.
Furthermore, even when considering only the false positive (red) points, VesselFusion produces line-like outputs that are not anatomically implausible, suggesting that the proposed method preserved vessel structures even in its false positive predictions.
\begin{figure}
    \centering
    \includegraphics[width=0.85\linewidth]{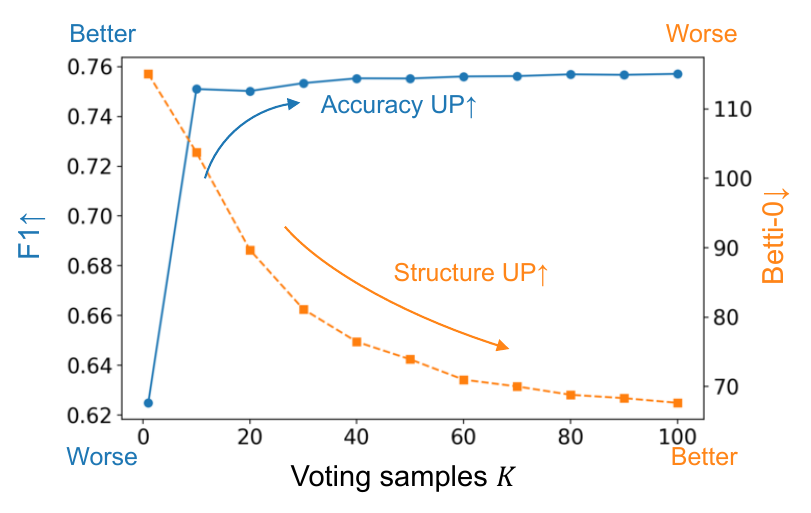}\\[-2mm]
    \caption{The effect of the number of samples $K$ in VesselFusion. The horizontal axis shows the number of samples $K$ used for voting, the left vertical axis shows the F1 score at $R=1$ (the higher the better), and the right vertical axis shows Betti-0 (the lower the better).}
    \label{fig:voting}
    \vspace{-10pt}
\end{figure}

\subsection{Ablation Study}
\label{sec:Ablation}
As shown in Table~\ref{tab:ablation_results}, we investigated the contribution of each component of VesselFusion: the coarse-to-fine coordinate representation (C2F) and the voting-based aggregation.
When introducing only C2F or only the voting scheme, all coordinate-based accuracy indicators improved compared with using neither component. In particular, C2F yielded a more pronounced gain in recall, whereas the voting scheme mainly boosted precision. Both components also reduced Betti-0, and the voting scheme additionally improved Betti-1, indicating enhanced robustness and structural consistency. Combining C2F with the voting scheme further boosted the F1-score and simultaneously improved both Betti-0 and Betti-1.

Fig.~\ref{fig:voting} shows the effect of the number of samples \(K\) used in the voting.  
As \(K\) increased, both coordinate accuracy and structural consistency improved: F1 rose while Betti numbers decreased.  
The F1-score saturated around $K=10$, suggesting that a relatively small number of samples is sufficient for accurate estimation, whereas Betti-0 continued to improve up to $K=60$, indicating that increasing the number of samples yields more stable estimation.

\section{Conclusion}
\label{sec:Conclusion}

We proposed VesselFusion, a diffusion-based method for vessel centerline extraction from 3D CT images.
By leveraging a coarse-to-fine coordinate representation and voting-based aggregation, VesselFusion enables stable training and accurate inference.
Experiments on a public CT dataset showed improved coordinate accuracy and anatomically plausible vessel structures over conventional methods, while computational cost related to voting-based aggregation remains a limitation for future work.



\subsubsection*{Compliance with ethical standards}\label{sec:ethics}
This study used the publicly available, anonymized ImageCAS~\cite{Zeng2023imagecas} dataset. 
No additional ethical approval was required under its license.

\subsubsection*{Acknowledgments}
\label{sec:acknowledgments}
This work was supported by SIP-JPJ012425, KAKEN JP24KJ1805, JP25K22846, and ASPIRE Grant Number JPMJAP2403.


\bibliographystyle{IEEEbib}
\bibliography{strings,refs}

\end{document}